\pdfoutput=1

\documentclass[11pt]{article}

\usepackage[final]{acl}

\usepackage{times}
\usepackage{latexsym}
\usepackage{mathtools}

\usepackage{algorithm}
\usepackage{algpseudocode}
\usepackage{amsmath}
\usepackage{float}
\usepackage{amssymb}
\usepackage{multirow}
\usepackage{xspace}
\usepackage{enumitem}
\usepackage{subfiles}
\usepackage{booktabs}

\usepackage[T1]{fontenc}

\usepackage[utf8]{inputenc}

\usepackage{microtype}

\usepackage{inconsolata}

\usepackage{graphicx}

\title{\name: Cross-Lingual Knowledge Transfer from High-Resource to Extreme Low-Resource Languages}

\author{Subhadip Maji \qquad Arnab Bhattacharya \\
        Department of Computer Science and Engineering, \\ 
        Indian Institute of Technology Kanpur \\
        \texttt{\{subhadip,arnabb\}@cse.iitk.ac.in}}

\setlist{nosep}
\newcommand{\name}{BhashaSetu\xspace}

\newcommand{\sm}[1]{}

\newcommand{\comment}[1]{}
\newcommand{\tabcaption}[1]{\caption{#1}\vspace*{-3mm}}

\begin{document}
\maketitle

\begin{abstract}
	Despite remarkable advances in natural language processing, developing effective systems for low-resource languages remains a formidable challenge, with performances typically lagging far behind high-resource counterparts due to data scarcity and insufficient linguistic resources. Cross-lingual knowledge transfer has emerged as a promising approach to address this challenge by leveraging resources from high-resource languages. In this paper, we investigate methods for transferring linguistic knowledge from high-resource languages to low-resource languages, where the number of labeled training instances is in hundreds. We focus on sentence-level and word-level tasks. We introduce a novel method, GETR (Graph-Enhanced Token Representation) for cross-lingual knowledge transfer along with two adopted baselines (a)~augmentation in hidden layers and (b)~token embedding transfer through token translation. Experimental results demonstrate that our GNN-based approach significantly outperforms existing multilingual and cross-lingual baseline methods, achieving 13 percentage point improvements on truly low-resource languages (Mizo, Khasi) for POS tagging, and 20 and 27 percentage point improvements in macro-F1 on simulated low-resource languages (Marathi, Bangla, Malayalam) across sentiment classification and NER tasks respectively. We also present a detailed analysis of the transfer mechanisms and identify key factors that contribute to successful knowledge transfer in this linguistic context. 
\end{abstract}

\section{Introduction}

Cross-lingual knowledge transfer has emerged as a crucial approach for improving natural language processing capabilities across different languages. Recent advances in multilingual model variants have demonstrated remarkable success in this domain by jointly training on multiple languages simultaneously, enabling zero-shot and few-shot learning capabilities. These models, such as XLM-R \cite{conneau2020unsupervised}, XLM-V \cite{xlm-v} and mmBERT \cite{mmbert2024}, learn shared representations across languages, thereby facilitating knowledge transfer from high-resource to low-resource languages. The success of these models largely stems from their ability to leverage massive multilingual corpora and transformer-based architectures \cite{vaswani2017attention}, which effectively capture cross-lingual patterns and relationships.

However, when dealing with extremely low-resource scenarios where target languages have very limited labeled data (e.g., only 100 training instances), even state-of-the-art multilingual models struggle  \cite{wu2020are, downey2024targeted, cassano2024knowledge} to generalize effectively. While parameter-efficient techniques like Adapter fine-tuning \cite{houlsby2019parameter} extreme and LoRA \cite{hu2022lora} reduce overfitting by updating fewer parameters, they still underperform in such extreme low-resource settings. Cross-lingual models like AdaMergeX \cite{zhao2025adamergex} also fail to capture sufficient linguistic nuances when number of target language examples is low.

Most languages worldwide have extremely limited digital resources -- India alone has at least hundreds of such languages, including several of its 22 official languages like Dogri, Bodo, Kashmiri and Santali. For practical evaluation in this work, we use comparatively higher-resource languages (Marathi, Bangla, Malayalam) as \emph{simulated} low-resource scenarios to enable rigorous testing with sufficient evaluation data. To validate our approach on \emph{truly} low-resource languages, we also evaluate on two real low-resource languages: Mizo and Khasi, which have only 502 and 507 annotated sentences respectively for POS tagging tasks.

 To address this challenge of extreme low-resource setting, we propose a comprehensive framework that intelligently transfers linguistic knowledge from high-resource to low-resource languages through one novel and two adopted complementary approaches. We name our approach \textbf{\name} after the words ``Bhasha'' and ``Setu'' that mean ``language'' and ``bridge'' respectively in most Indian languages, highlighting its role in bridging languages.

Our approach is as follows. First, we introduce an adopted baseline, Hidden Augmentation Layers (HAL) that create mixed representations in the hidden space, allowing controlled knowledge transfer while preserving the target language's distinctive features. Second, we develop another adopted baseline, a token embedding transfer mechanism that leverages translation-based mappings to initialize low-resource language embeddings effectively. Finally, we propose a novel Graph-Enhanced Token Representation (GETR) approach that uses Graph Neural Networks \cite{zhou2020graph, kipf2017semi, velivckovic2018graph} to enable dynamic knowledge sharing between languages at the token level, thereby capturing complex cross-lingual relationships through graph-based message passing.
In short, our contributions are:

\begin{enumerate}
	
	\item We propose a comprehensive framework, \textsc{\name}, for cross-lingual knowledge
	transfer in extreme low-resource scenarios, comprising two adopted baselines such as hidden augmentation layer (HAL) and token
	embedding transfer (TET), and a novel graph-enhanced token representation (GETR) with
	GNNs (Sec.~\ref{sec:methodology}).
	
	\item We conduct extensive experiments across multiple NLP tasks and language pairs spanning multiple languages, demonstrating the versatility and robustness of our approach. Experimental results on POS tagging task low-resource languages Mizo and Khasi using high-resource languages Hindi and English demonstrate that our novel GNN-based approach significantly outperforms existing methods, achieving 13 percentage points improvement respectively in macro-F1 score compared to traditional multilingual and cross-lingual baselines while requiring only 100 training instances in the low-resource language (Sec.~\ref{sec:results}). 
	
	\item We provide systematic analysis of the impact of various factors on
	cross-lingual knowledge transfer, including mixing coefficient,
	architectural depth and dataset size ratios between
	languages (Sec. \ref{sec:results}). 
\end{enumerate}

\section{Related Work} 
\label{sec:related_work}

Cross-lingual transfer learning has advanced significantly with multilingual pre-trained models such as XLM-R \cite{conneau2020unsupervised} and mmBERT \cite{mmbert2024}. While effective, these approaches require substantial multilingual training data, limiting their applicability in extreme low-resource settings. Recent parameter-efficient fine-tuning methods like LoRA \cite{hu2022lora}, AdaMergeX \cite{zhao2025adamergex}, and SALT \cite{lee-etal-2025-semantic-aware} reduce overfitting risks but struggle with extremely limited target data. 

Data augmentation techniques in hidden spaces, including mixup \cite{zhang2018mixup, verma2019manifold} and their NLP adaptations \cite{chen2020mixtext, sun2020mixup}, have proven valuable for low-resource scenarios and are comprehensively surveyed by \citet{feng2021survey}. Token-level transfer approaches like trans-tokenization \cite{transtokenization} enable cross-lingual embedding transfer without requiring parallel data, addressing a critical challenge for low-resource languages.

Graph-based cross-lingual methods such as Heterogeneous GNNs \cite{wang2021cross} depend on external semantic parsers and operate solely at the GNN level, without integrating graph knowledge into transformer models. Colexification-based multilingual graphs \cite{liu2023crosslingual} construct graphs from colexification relations rather than token interactions, and similarly do not infuse graph information into transformers. While recent work has employed graph-based transformers with UCCA semantic graphs \cite{wan2024exploring}, such approaches require pre-trained semantic parsers that are typically unavailable for low-resource Indian languages. In contrast, our GETR method constructs token-level graphs directly from training data and uniquely integrates GNN-based token interactions within the transformer, enabling dynamic, fine-grained cross-lingual knowledge sharing without external linguistic resources.

\section{Methodology}
\label{sec:methodology}


This section presents three approaches for cross-lingual knowledge transfer: (a)~\textbf{Augmentation in Hidden Layers (HAL)}, an adaptation of existing layer-wise mixing strategies; (b)~\textbf{Token Embedding Transfer through Translation (TET)}, a refinement of token embedding transfer through translation dictionaries; and (c)~\textbf{Graph-Enhanced Token Representation (GETR)}, a novel approach leveraging Graph Neural Networks to dynamically share token embeddings across languages at hidden layers. While HAL and TET build upon established techniques from prior work, both are adapted specifically for the cross-lingual setting with modifications tailored to leverage high-resource language knowledge for low-resource language tasks. In contrast, GETR introduces a fundamentally new mechanism that constructs dynamic token-level graphs to enable fine-grained cross-lingual knowledge sharing through neighborhood aggregation in GNNS. Before delving into the technical details of these approaches, we first formally define the problem statement.


%

\paragraph{Problem Statement:} Let us formally define our notation for cross-lingual knowledge transfer. For a high-resource language, we denote the dataset of textual instances as $\mathbf{X_H} = \{x_1, x_2, \ldots, x_{N_H}\}$, where each $x_i$ represents an individual text instance (e.g., a sentence). The corresponding task-specific outputs are represented as $Y_H = \{y_1, y_2, \ldots, y_{N_H}\}$. Here, $N_H$ represents the total number of instances in the high-resource dataset, typically in the order of thousands or more. Similarly, we denote the low-resource language dataset as $\mathbf{X_L}$ and its corresponding outputs as $Y_L$, where $|\mathbf{X_L}| = N_L \ll N_H$, with $N_L$ being extremely small (approximately 100 instances). This extreme data scarcity in the low-resource setting presents the core challenge in our task.

We define the combined dataset as $\mathbf{X} = \{\mathbf{X_H} \cup
\mathbf{X_L}\}$
and $Y = \{Y_H \cup Y_L\}$.
Our objective is to learn a
model $\mathbf{M}: \mathbf{X} \to Y$ that maps input text instances from
either or both $\mathbf{X_H}$ and $\mathbf{X_L}$ to their respective outputs,
while effectively leveraging the high-resource language data to compensate
for the limited low-resource samples. The output space $Y$ can correspond to
any encoder-based task, with two common task variants.
The first is for sentence-level tasks (such as sentiment analysis) where $y_i$ is one of $c$ classes.
The second is for sequence-labeling tasks (such
as NER): $y_i = [y_{i_1}, y_{i_2}, \ldots, y_{i_T}]$,
where $T$ is the sequence length and each token-level label $y_{it} \in
\mathcal{Y}_{tags}$ represents a class (such as an NER tag).

Despite the different output structures, the core challenge of effective cross-lingual knowledge transfer remains consistent across tasks, allowing us to apply the same methodological approaches with task-specific adaptations. We next describe the three methods.

\subsection{HAL}

Hidden layer augmentation has emerged as a prevalent technique for generating synthetic training data in the latent space when working with textual inputs \cite{zhang2018mixup, verma2019manifold}. While this approach has been successfully applied for domain adaptation within the same language \cite{mtd}, its application to cross-lingual knowledge transfer, particularly from high-resource to low-resource languages, represents a novel direction. This method is particularly versatile as it can be applied to any high-resource and low-resource language pair, regardless of their script similarities.

Let $\mathbf{E_M}: \mathbf{X} \to \mathbf{H}$ denote the encoder component of the model $\mathbf{M}$ that maps each input text $x_i$ to its final encoded \texttt{CLS} representation $h^{\text{CLS}}i$. We propose a hidden augmentation mechanism that fuses knowledge from high-resource and low-resource languages through a weighted combination in the latent space. Formally, we generate new training pairs $A_i = (h^{\text{CLS}}_{A_i}, y_{A_i})$ as follows: 
\begin{align} 
 	h^{\text{CLS}}_{A_i} &= \alpha \cdot h^{\text{CLS}}_{H_i} + (1-\alpha) \cdot h^{\text{CLS}}_{L_i} \\
 	y_{A_i} &=
 	\begin{cases}
 		\alpha \cdot y_{H_i} + (1{-}\alpha) \cdot y_{L_i}, \\[2pt]
 		\left(\alpha \cdot y_{H_i,t} + (1{-}\alpha) \cdot y_{L_i,t}\right)_{t=1}^T
 	\end{cases}
 	 \label{eq:hidden_aug_h} 
\end{align} 
where $\alpha \in [0,1]$ is a mixing coefficient that controls the contribution of each language. This coefficient can be either fixed through training or randomly sampled per iteration. For sentence tasks with $c$ classes, $y_{H_i}, y_{L_i} \in \mathbb{R}^c$ are typically one-hot encoded vectors, while for sequence tasks, $y_{H_i,t}, y_{L_i,t} \in \mathbb{R}^{|\mathcal{Y}_{tags}|}$ represent the tag distribution at position $t$.

Empirically, $\alpha$ values between 0.1 and 0.4 yield optimal results, as they maintain the primary characteristics of the low-resource language while supplementing it with knowledge from the high-resource language. Since the augmentation produces soft labels, we employ KL-divergence loss \cite{cui2023decoupled} instead of standard cross-entropy loss \cite{zhong2023crossentropy} for soft labels and cross-entropy for hard labels during training. This framework can be further extended by adding multiple transformer layers above $E_M$ and performing augmentation at each layer's \texttt{CLS} output, thus enabling hierarchical knowledge fusion.






\subsection{TET}
\label{sec:tet}
Traditional approaches often initialize token embeddings for low-resource languages randomly, which can lead to suboptimal performance, especially when training data is scarce. We propose an initialization strategy that leverages token embeddings from a high-resource language through translation mapping \cite{transtokenization}. This approach provides a more informed starting point for the embedding matrix of the low-resource language, thereby enabling effective fine-tuning even with limited training samples. The core idea is to initialize the token embeddings of the low-resource language using the semantic information captured in the pre-trained embeddings of their translated counterparts in the high-resource language. While this method assumes the availability of word-level translations for the training data of the low-resource language, it does not require any pre-trained models or large corpora in the low-resource language. In our experiments, we used pymultidictionary \cite{PyMultiDictionary} primarily to make our translation process seamless, faster and automated for the languages in our study, followed by manual verification of the translations to ensure accuracy. For extremely low-resource languages without dictionary support, we recommend manually translating the limited training vocabulary (which is manageable given the small dataset size of $\sim$100 instances).

\begin{algorithm}[tbp]
	{\small
		\caption{Token Embedding Transfer through Translation (TET)}
		\label{alg:token_transfer}
		\begin{algorithmic}[1]
			\State $V_L \gets$ Set of unique words from LRL corpus
			\ForAll{$w_l \in V_L$}  \Comment{For each LRL word}
			\State $w_h \gets \text{TranslateToHRL}(w_l)$ 
			\State $T_h \gets \text{HRLTokenize}(w_h)$ 
			\State $T_l \gets \text{LRLTokenize}(w_l)$ 
			\State $E_h \gets \{\text{GetPretrainedEmbeddings}(t) | t \in T_h\}$ \Comment{HRL token embeddings}
			\State $e_{\text{avg}} \gets \text{Mean}(E_h)$
			\ForAll{$t_l \in T_l$}  \Comment{For each LRL token}
			\State $P_{t_l} \gets \emptyset$ \Comment{Initialize projected embeddings set}
			\ForAll{$w' \in V_L$} \Comment{Check all LRL words}
			\If{$t_l \in \text{LRLTokenize}(w')$} 
			\State $P_{t_l} \gets P_{t_l} \cup \{e_{\text{avg}}\}$
			\EndIf
			\EndFor
			\State $E_l[t_l] \gets \text{Mean}(P_{t_l})$ \Comment{Final embedding for LRL token}
			\EndFor
			\EndFor
			\State \Return $E_l$ \Comment{Dictionary of LRL token embeddings}
		\end{algorithmic}
	}
\end{algorithm}

Algorithm~\ref{alg:token_transfer} details our systematic process for
transferring token embeddings from a high-resource language (e.g., English)
to a low-resource language (e.g., Marathi). To illustrate this process,
consider transferring embeddings for the Marathi word "\=antarbh\=a\d{s}ika"
meaning "cross-lingual" in English. The word would be translated to English as "cross-lingual", which might be tokenized as "cross" + "lingual" in English. The word, "\=antarbh\=a\d{s}ika" would be tokenized in Marathi,
potentially splitting it into subword tokens like "\=antar" + "bh\=a\d{s}ika".
The pre-trained embeddings for these English
tokens are retrieved and averaged. For each Marathi token, we collect all
instances where it appears across different words in the Marathi corpus. For
example, the token "bh\=a\d{s}ika" might also appear in words like
"bahubh\=a\d{s}ika" (meaning "multi-lingual"). Finally, we average all
corresponding English embedding projections to create the final embedding for
each Marathi token. While we show transliterated examples here for
clarity, in our actual experiments we used the original scripts for all
languages.

\subsection{GETR}
We propose a novel approach leveraging Graph Neural Networks (GNN) \cite{zhou2020graph} to enable dynamic knowledge sharing between high-resource and low-resource languages at the token level. For each batch of mixed-language inputs, we construct an undirected graph $G = (T, C)$, where $T = \{t_1, t_2, \ldots, t_{N_k}\}$ represents the set of $N$ unique tokens in batch $k$. The edge set $C$ captures sequential relationships between tokens, defined as $C \subseteq \{{t_{ij}, t_{i(j+1)}} | t_{ij}, t_{i(j+1)} \in T\}$, where tokens $t_{i1}, t_{i2}, \ldots, t_{in}$ form sentence $s_i$.

\begin{figure}[tbp]
	\centering
	\includegraphics[width=0.75\linewidth]{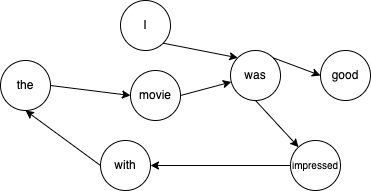}
    \vspace*{-1mm}
	\caption{Graphical representation of tokens of two sentences in a batch: ``The movie was good'' and ``I was impressed with the movie''.}
	\label{fig:node_graph}
	\vspace*{-4mm}
\end{figure}

To illustrate the mechanism, consider two sentences: "The movie was good" from a high-resource language and "I was impressed with the movie" from a low-resource language. As shown in Figure \ref{fig:node_graph}, tokens are represented as nodes with edges connecting consecutive tokens within each sentence. When computing the representation for shared tokens (e.g., "was"), the model incorporates contextual information from both language environments. This allows the \texttt{CLS} embedding of the low-resource sentence to benefit from the high-resource language's token representations through neighborhood aggregation.

\begin{figure}[t]
	\centering
	\includegraphics[width=0.75\linewidth]{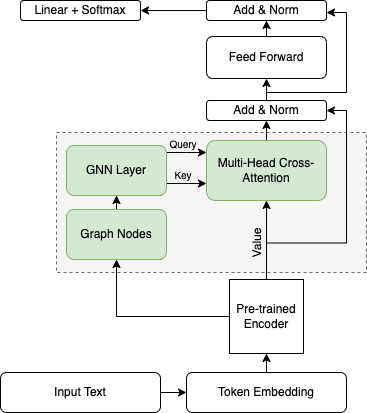}
	\vspace*{-1mm}
	\caption{BERT encoder architecture incorporating the GNN layer for cross-lingual knowledge transfer.}
	\label{fig:gnn_arch}
	\vspace*{-4mm}
\end{figure}

Given the encoder output $\mathbf{H}\in \mathbb{R}^{B \times S \times D}$
(where $B$, $S$, and $D$ denote batch size, sequence length, and embedding
dimension respectively), we reshape it to $\mathbf{H'} \in \mathbb{R}^{L \times D}$ ($L = B \times S$) for GNN processing. We employ either GCN \cite{kipf2017semi} or GAT
\cite{velivckovic2018graph} layers with an adjacency matrix $\mathbf{A} \in
\{0, 1\}^{L \times L}$ that captures token relationships such as
$\mathbf{A}_{ij} = 1$ if $l_i$ and $l_j$ are consecutive tokens in a sentence.
Notably, we construct $\mathbf{A}$
using the flattened dimension $L$ rather than unique tokens, allowing for token
repetition which makes the array multiplication simpler and straight-forward.
The GNN output is then reshaped to generate query $\mathbf{Q}$ and key
$\mathbf{K}$ matrices for the subsequent transformer layer:
\begin{align}
	\begin{aligned}
		\mathbf{H'} &= \text{Reshape}(\mathbf{H}) \in \mathbb{R}^{L \times D} \\
		\mathbf{H'_G} &= \text{GNN}(\mathbf{H'}) \\
		\mathbf{H_G} &= \text{Reshape}(\mathbf{H'_G}) \in \mathbb{R}^{B \times S \times D} \\
		\mathbf{Q} &=  \mathbf{H_G} \times \mathbf{W_q} \\
		\mathbf{K} &=  \mathbf{H_G} \times \mathbf{W_k}
	\end{aligned}
\end{align}
where $\mathbf{W_q} \in \mathbb{R}^{D \times D'}$ and $\mathbf{W_k} \in \mathbb{R}^{D \times D'}$ are query and key weight matrices respectively. The subsequent transformer operations remain unchanged, following the standard sequence of cross-attention, feed-forward networks, layer normalization, and residual connections.
The value
$\mathbf{V}$ matrix maintains its original computation path:
\begin{align}
	\begin{aligned}
		\mathbf{V} &=  \mathbf{H} \times \mathbf{W_v} \\
	\end{aligned}
\end{align}
where $\mathbf{W_v} \in \mathbb{R}^{D \times D'}$ is the value weight matrix. Once $\mathbf{Q}$, $\mathbf{K}$ and $\mathbf{V}$ are computed, the rest of the transformer encoder \cite{vaswani2017attention} block is unchanged, i.e., cross-attention block followed by feed-forward, layer normalization and residual connection. Figure \ref{fig:gnn_arch} illustrates our modified BERT architecture with GNN layers (gray shaded area). Multiple GNN layers can be stacked sequentially to enable deeper cross-lingual knowledge transfer.

%
\paragraph{Strategic Batch Formation for Graph Construction:} We propose a batch
formation strategy that balances high-resource and low-resource instances while
maximizing token overlap between languages. For every batch of size $B$, we
ensure exactly $B/2$ instances from each language domain. Our construction
alternates between low-resource and high-resource anchors: we first select a
random low-resource instance, then add $(n/2-1)$ neighbors from low-resource
language and $n/2$ from high-resource language based on maximum token overlap.
These $n$ instances are removed from the available pool to prevent repetition
within an epoch. We then select a high-resource anchor and repeat the process,
and continue this alternation until the batch is filled. 
To improve robustness, 70\% of the batches follow this strategic formation
while the remaining 30\% maintain an equal language distribution that selects
instances randomly. This prevents over-reliance on specific token patterns
while preserving structured knowledge transfer. The process continues across
epochs until all low-resource instances are utilized.

During inference, we apply the same principle using training data to form neighborhoods for test instances based on token overlap. This balanced batch construction creates our token interaction graph $G = (T, C)$, enabling effective cross-lingual token relationships without requiring pre-trained resources for the low-resource language.

\paragraph{Cross-Script Edge Construction:} When sentences in a batch use different scripts, we establish edges between tokens based on semantic equivalence. We utilize the same translation mechanism described in our TET method (Sec.~\ref{sec:tet}) -- a combination of dictionary lookup via pymultidictionary \cite{PyMultiDictionary} followed by manual verification.

When two sentences appear in the same batch (e.g., $s_1$ from English and $s_2$ from Marathi), and the $i$-th word of $s_1$ ($w_i^E$) conveys similar meaning as the $j$-th word of $s_2$ ($w_j^M$) based on our translation dictionary, connections are established between their tokens. If these words are each represented by single tokens, one edge connects them. If tokenizers split these words into multiple sub-tokens, then every token of $w_i^E$ will be connected to every token of $w_j^M$.

For example, if ``antarbhasika'' (in Marathi) and ``cross-lingual'' (in English) are present in different sentences of a batch, and they are tokenized as ``antar'' + ``bhasika'' and ``cross'' + ``lingual'' respectively, edges will be established between all pairs: ``antar''-``cross'', ``antar''-``lingual'', ``bhasika''-``cross'' and ``bhasika''-``lingual''. While some connections may be noisy, this approach works because: (1) some connections are semantically appropriate (e.g., ``antar''-``cross'' and ``bhasika''-``lingual''), and (2) such cross-token scenarios occur infrequently ($<20\%$ of connections in a typical batch). During training, the model learns to adjust weights for these connections based on their usefulness for the objective function.

\section{Experiments and Results}
\label{sec:results}

\subsection{Dataset}

Our experiments evaluate cross-lingual knowledge transfer across multiple languages and tasks.

For sentiment classification, we employ two high-resource languages: Hindi \cite{sid573_hindi_sentiment, sawant_hindi_sentiment} and English \cite{akanksha_sentiment_dataset}, each with 12,000 labeled instances. We use two low-resource target languages: Marathi \cite{marathi_dataset}, which shares the Devanagari script with Hindi, and Bangla (Bengali) \cite{sazzed2019sentiment}, a language close to Hindi but with its own script. For the Marathi and Bangla datasets, we \emph{simulate} extreme low-resource scenarios with the following random splits: 100 training instances, 1,500 validation instances, and 2,000 test instances. All sentiment classification datasets contain binary labels (positive and negative) with balanced class distributions.

For Named Entity Recognition (NER), we maintain English and Hindi as high-resource languages, with 12,000 training instances (17 entity tags) and 12,084 training instances (13 entity tags) respectively \cite{namanj27_ner_2022, murthy-etal-2022-hiner}. For low-resource languages, we use Marathi \cite{patil2022l3cube} and Malayalam \cite{mhaske2022naamapadam} (which uses a completely different script) with 100 training, 1,500 validation, and 2,000 test instances (14 and 7 unique entity tags for Marathi and Malayalam respectively) created by random sampling.

For the Part-of-Speech (POS) tagging task, we compiled two \emph{truly} low-resource language datasets \cite{Ghosh2025}: Mizo (502 sentences) and Khasi (507 sentences). Each sentence is fully annotated with POS tags. From both datasets, 100 sentences were randomly selected as the training set and 100 as the validation set; the remainder 302 sentences for Mizo and 307 for Khasi formed the test set. High-resource POS datasets comprise 12,000 labeled sentences each for English \cite{Silveira2021} and Hindi \cite{BhatYear}.

\subsection{Implementation Details}

Following our extreme low-resource assumption, we first used a tinyBERT \cite{jiao2019tinybert} architecture (without pre-training) to train models from scratch using only 100 labeled instances of each low-resource language, including training new tokenizers. Additionally, we trained two classical ML models such as XGBoost for sentiment classification and CRF for NER and POS tagging tasks. We established several baselines for comparison: (1)~\emph{Joint Training (JT)}, which trains on high and low resource languages simultaneously, similar to multilingual models; (2)~\emph{JT-HRLAdapLRL}, which sequentially fine-tunes on high-resource data followed by adapter-based fine-tuning on low-resource data; (3)~\emph{XLMFT-HRLAdapLRL}, which applies the same sequential approach but starts from XLM-R \cite{conneau2020unsupervised}; (4)~\emph{LoRA} \cite{hu2022lora}, which updates low-rank decomposition matrices instead of full weights; and (5)~\emph{AdaMergeX} \cite{zhao2025adamergex}, which combines multiple adaptation methods. For adapter fine-tuning, we used a reduction factor of 16, increasing total parameters by only 1\%. For LoRA and AdaMergeX, we followed the recommended parameter settings for encoder tasks from their original papers \cite{hu2022lora,zhao2025adamergex}.

For our high-resource languages, we utilized \texttt{l3cube-pune/hindi-albert} \cite{joshi2022l3cubehind} for Hindi and \texttt{albert/albert-base-v2} \cite{albert} for English across both tasks. All experiments were conducted on an Amazon EC2 \texttt{p4de.24xlarge} instance with 8 NVIDIA A100 GPUs (80 GB each), using batch sizes of 128 for most approaches (8 for Scratch Training due to low number of data points, and 120 for GETR methods to accommodate graph construction using 9 neighbors per instance). We employed AdamW with learning rates between 3e-5 and 3e-7 for pre-trained models, and 3e-4 for Scratch Training with TET. Models were trained for 50 epochs with best checkpoints selected via validation loss.

For a fairer comparison between methods, we carefully balanced parameter counts across all models. Since GETR adds additional GNN layers to the architecture, we removed transformer layers from the pre-trained model to maintain comparable model size. For example, in GETR-GAT, we removed the last 3 transformer encoder layers and added 2 GAT layers, resulting in a parameter count (237,558,024) nearly identical to the Joint Training model (237,557,762). All experiments used the original scripts of the respective languages rather than transliteration. For baseline models, HAL and GETR approaches, we leveraged pre-trained tokenizers from high-resource languages, augmenting them with new tokens from low-resource languages. These newly added tokens were randomly initialized, allowing the model to learn appropriate representations during training


All reported results are evaluated on carefully selected test sets to ensure that there is no overlap with training data (Table~\ref{tab:all_results}). As expected, with such limited data and no pre-trained knowledge, Scratch Training models failed to learn meaningful patterns, defaulting to macro-F1 scores of 0.33-0.38 for sentiment classification, and 0.03-0.17 for NER and 0.02-0.16 for POS tagging.

\subsection{Results on Sentiment Classification Task}

Using English as the high-resource language, the baseline models achieve macro-F1 scores ranging from 0.53 to 0.55 for Marathi, with AdaMergeX performing best among them. Our proposed HAL method with $\alpha=0.2$ and two layers shows improvement (0.63 with TET), but the GETR-GAT approaches demonstrate substantially greater gains, with GETR-GAT+HAL achieving the best performance (0.75 macro-F1), representing a 20 percentage point improvement over the best baseline. For Bangla as the low-resource language, baseline models achieve macro-F1 scores of 0.63, while GETR-GAT+HAL+TET delivers the best performance at 0.75, a 12 percentage point improvement.

With Hindi as the high-resource language, baseline performance improves significantly (up to 0.76 for AdaMergeX with Marathi), highlighting the benefit of script similarity. When Hindi is used as HRL and Marathi as LRL, TET is not required as they share the Devanagari script, ensuring that Marathi tokens already have pre-trained embeddings from the Hindi model. This explains the absence of TET-based results for this language pair in Table \ref{tab:all_results}. Our HAL approach further boosts performance (0.80 macro-F1), while GETR-GAT+HAL achieves the highest score for Marathi (0.87 macro-F1), an 11 percentage point improvement over the best baseline. For Hindi-Bangla, GETR-GAT+HAL+TET reaches 0.81 macro-F1, outperforming the best baseline (AdaMergeX at 0.69) by 12 percentage points.

\begin{table*}[htbp]
	\centering
	\resizebox{0.99\textwidth}{!}{
		\begin{tabular}{llllllll}
			\toprule
			\multirow{2}{*}{\textbf{HRL}} & \multirow{2}{*}{\textbf{Method}} & \multicolumn{2}{c}{\textbf{Sentiment Classification}} & \multicolumn{2}{c}{\textbf{NER}} & \multicolumn{2}{c}{\textbf{POS Tagging}} \\
			\cmidrule(lr){3-4} \cmidrule(lr){5-6} \cmidrule(lr){7-8}
			& & \textbf{Marathi} & \textbf{Bangla} & \textbf{Marathi} & \textbf{Malayalam} & \textbf{Mizo} & \textbf{Khasi} \\
			\midrule
			- & Scratch Training (tinyBERT) & 0.33 $\pm$0.000  & 0.33 $\pm$0.000  & 0.03 $\pm$0.073  & 0.03 $\pm$0.073  & 0.02 $\pm$0.061 & 0.04 $\pm$0.084 \\
            - & Scratch Training (XGBoost) & 0.38 $\pm$0.001  & 0.36 $\pm$0.001  & - & -  & - & - \\
            - & Scratch Training (CRF) & -  & -  & 0.17 $\pm$0.004 & 0.15 $\pm$0.002  & 0.16 $\pm$0.002 & 0.14 $\pm$0.003 \\
			\midrule
			\multirow{11}{*}{English} & Joint Training & 0.53 $\pm$0.002  & 0.63 $\pm$0.001  & 0.29 $\pm$0.001  & 0.26 $\pm$0.002  & 0.75 $\pm$0.010 & 0.71 $\pm$0.012 \\
			& Adapter Finetuning & 0.53 $\pm$0.001  & 0.63 $\pm$0.002  & 0.29 $\pm$0.001  & 0.26 $\pm$0.001  & 0.76 $\pm$0.008 & 0.72 $\pm$0.008 \\
			& XLMFT-HRLAdapLRL & 0.53 $\pm$0.001  & 0.63 $\pm$0.002  & 0.29 $\pm$0.002  & 0.26 $\pm$0.001  & 0.76 $\pm$0.009 & 0.72 $\pm$0.011 \\
			& LoRA & 0.54 $\pm$0.001  & 0.63 $\pm$0.001  & 0.29 $\pm$0.001  & 0.27 $\pm$0.001  & 0.78 $\pm$0.007 & 0.73 $\pm$0.008 \\
			& AdaMergeX & 0.55 $\pm$0.001  & 0.63 $\pm$0.001  & 0.29 $\pm$0.001  & 0.28 $\pm$0.002  & 0.79 $\pm$0.010 & 0.75 $\pm$0.007 \\
			& HAL & 0.60 $\pm$0.001  & 0.64 $\pm$0.001  & 0.32 $\pm$0.001  & 0.30 $\pm$0.003  & 0.85 $\pm$0.009 & 0.79 $\pm$0.008 \\
			& HAL + TET & 0.63 $\pm$0.001  & 0.65 $\pm$0.003  & 0.33 $\pm$0.001  & 0.31 $\pm$0.002  & - & - \\
			& GETR-GAT & 0.73 $\pm$0.001  & 0.72 $\pm$0.001  &  \textbf{0.40} $\pm$0.001 & 0.46 $\pm$0.001  & 0.91 $\pm$0.007 & 0.86 $\pm$0.007 \\
			& GETR-GAT + TET & 0.74 $\pm$0.001  & 0.73 $\pm$0.002  &  \textbf{0.40} $\pm$0.001 & 0.47 $\pm$0.003  & - & - \\
			& GETR-GAT + HAL & \textbf{0.75} $\pm$0.001  & 0.74 $\pm$0.001  & \textbf{0.40} $\pm$0.001 & 0.51 $\pm$0.002  & \textbf{0.92} $\pm$0.006 & \textbf{0.88} $\pm$0.006 \\
			& GETR-GAT + HAL + TET & 0.74 $\pm$0.001  & \textbf{0.75} $\pm$0.001  & \textbf{0.40} $\pm$0.001  & \textbf{0.52} $\pm$0.001  & - & - \\
			\midrule
			\multirow{11}{*}{Hindi} & Joint Training & 0.75 $\pm$0.004  & 0.67 $\pm$0.003  & 0.35 $\pm$0.002  & 0.28 $\pm$0.002  & 0.75 $\pm$0.009 & 0.71 $\pm$0.007 \\
			& Adapter Finetuning & 0.74 $\pm$0.002  & 0.67 $\pm$0.002  & 0.34 $\pm$0.003  & 0.28 $\pm$0.002  & 0.76 $\pm$0.009 & 0.71 $\pm$0.008 \\
			& XLMFT-HRLAdapLRL & 0.74 $\pm$0.002  & 0.67 $\pm$0.002  & 0.34 $\pm$0.002  & 0.28 $\pm$0.002  & 0.76 $\pm$0.008 & 0.72 $\pm$0.010 \\
			& LoRA & 0.75 $\pm$0.001  & 0.68 $\pm$0.001  & 0.35 $\pm$0.002  & 0.28 $\pm$0.001  & 0.77 $\pm$0.006 & 0.73 $\pm$0.009 \\
			& AdaMergeX & 0.76 $\pm$0.001  & 0.69 $\pm$0.001  & 0.36 $\pm$0.001  & 0.28 $\pm$0.001  & 0.78 $\pm$0.008 & 0.74 $\pm$0.007 \\
			& HAL & 0.80 $\pm$0.005 & 0.72 $\pm$0.004  & 0.38 $\pm$0.001  & 0.32 $\pm$0.003  & 0.83 $\pm$0.006 & 0.77 $\pm$0.009 \\
			& HAL + TET & - & 0.73 $\pm$0.002  & - & 0.32 $\pm$0.002  & 0.83 $\pm$0.005 & 0.77 $\pm$0.006 \\
			& GETR-GAT & 0.85 $\pm$0.001  & 0.79 $\pm$0.002  & \textbf{0.44} $\pm$0.001 & 0.48 $\pm$0.001  & 0.88 $\pm$0.008 & 0.82 $\pm$0.006 \\
			& GETR-GAT + TET & - & 0.80 $\pm$0.001  & - & 0.49 $\pm$0.003  & 0.88 $\pm$0.007 & 0.82 $\pm$0.006 \\
			& GETR-GAT + HAL & \textbf{0.87} $\pm$0.001  & 0.80 $\pm$0.003  & \textbf{0.44} $\pm$0.001  & 0.53 $\pm$0.002  & 0.88 $\pm$0.007 & \textbf{0.83} $\pm$0.007 \\
			& GETR-GAT + HAL + TET & - & \textbf{0.81} $\pm$0.002  & - & \textbf{0.55} $\pm$0.001  & \textbf{0.89} $\pm$0.008 & \textbf{0.83} $\pm$0.009 \\
			\bottomrule
		\end{tabular}
	}
    \vspace*{-1mm}
	\caption{Performance (Macro-F1 score) comparison of different training approaches on Sentiment Classification, NER, and POS Tagging tasks when Hindi and English are considered as HRL and Marathi, Bangla, Malayalam, Mizo, and Khasi as LRL. The mean and standard deviation numbers are reported based on 5 independent runs. \textbf{Bold} indicates results that are better with statistical significance ($p < 0.005$).}
	\label{tab:all_results}
    \vspace*{-3mm}
\end{table*}





\subsection{Results on NER Task}

We extended our evaluation to NER for Malayalam and Marathi. Baseline models achieve macro-F1 scores of 0.26--0.28 regardless of high-resource language choice (English or Hindi). GETR-GAT+HAL+TET substantially outperforms baselines: for Malayalam, reaching 0.52 with English (24 pp improvement) and 0.55 with Hindi (27 pp improvement); for Marathi, achieving 0.40 with English (11 pp above best baseline) and 0.44 with Hindi (8 pp above best baseline).

\subsection{Results on POS Tagging Task}

We evaluate Part-of-Speech tagging on two low-resource languages, Mizo and Khasi. Baseline performance is consistent across both languages and high-resource language choices at 0.71--0.75. GETR-GAT+HAL+TET delivers strong improvements: for Mizo, reaching 0.92 with English and 0.88 with Hindi; for Khasi, achieving 0.88 with English and 0.83 with Hindi, demonstrating the method's effectiveness on truly low-resource languages with minimal digital presence.


These consistent improvements across different tasks and language families (Indo-Aryan and Dravidian) demonstrate that our GETR approach effectively transfers knowledge regardless of task type or target language. GETR's superior performance can be attributed to its ability to create dynamic, contextualized connections between tokens across languages, enabling more effective knowledge transfer at a granular level. Unlike static approaches, GETR allows low-resource language tokens to directly incorporate relevant semantic information from high-resource contexts through the graph structure, creating richer representations that better capture cross-lingual patterns. This transfer mechanism operates efficiently through the transformer's multi-head attention, where Q and K matrices capture the graph-based knowledge of tokens while preserving the original value computations, allowing cross-lingual information to propagate throughout the network. We observed that when using more complex approaches like HAL or GETR, TET's contribution diminishes. We implemented additional baselines including HAL-LRL (augmentation within low-resource language only) and other three XLM-R finetuning variants, all performing comparably to Joint Training. We also evaluated GETR-GCN, but GETR-GAT consistently outperformed it due to GAT's adaptive edge weighting versus GCN's equal weighting of connections. Complete results for these experiments appear in Table~\ref{ap:all_results} (in Appendix).

\subsection{Cost and Environmental Impact}

GETR-GAT incurs modest computational overhead compared to Joint Training while delivering substantial performance gains. Training on AWS \texttt{p4de.24xlarge} instances requires approximately 11\% additional time per epoch ($\approx$ 50 minutes vs. $\approx$ 45 minutes for Joint Training) and 8\% more peak GPU memory ($\approx$ 41 GB vs. $\approx$ 38 GB) due to graph neural network computations and neighborhood construction. Over 50 training epochs, the total training time increases from 37.5 hours to 41.7 hours. Inference is approximately 6.3\% slower per sample (0.0101 vs. 0.0095 seconds on p3.2xlarge V100 GPUs). The increased energy consumption ($\approx$ 2.2 kWh vs. $\approx$ 2.0 kWh per full run) results in approximately 10\% higher $\text{CO}_2$ emissions ( $\approx$ 0.99 kg $\text{CO}_2$ vs. $\approx$ 0.90 kg $\text{CO}_2$), corresponding to roughly 0.09 kg additional $\text{CO}_2$ per model training. These modest computational costs are justified by the significant performance improvements as shown in Table~\ref{tab:all_results}. Detailed report is presented in Table~\ref{tab:computational_impact} (in Appendix).

\subsection{Ablation Studies}

To evaluate the robustness of our approach and demonstrate its advantage over baseline methods, we compared \name (our best-performing GETR-GAT+HAL configuration) with AdaMergeX \cite{zhao2025adamergex} across varying dataset sizes for NER with Hindi as HRL and Marathi as LRL (Table~\ref{tab:size_impact}). The results reveal two critical insights. First, with extremely limited low-resource data (10-50 instances), AdaMergeX achieves modest performance (0.05-0.17 F1), while \name demonstrates substantially better results even with minimal data, achieving 0.11 F1 with just 10 LRL instances and 0.34 F1 with 50 instances -- a 17 percentage points improvement over AdaMergeX at these data scales. The fixed HRL size (12,000) experiment shows \name's consistent advantage
across all LRL sizes, with improvements of 9-17 percentage points, though the
relative gap narrows as low-resource data increases. 

\begin{table}[tb]
	\centering
	\resizebox{0.9\columnwidth}{!}{
		\begin{tabular}{rrcc}
			\toprule
			\multirow{2}{*}{\textbf{LRL Size}} & \multirow{2}{*}{\textbf{HRL Size}} & \multicolumn{2}{c}{\textbf{Macro F1}} \\
			\cmidrule(lr){3-4}
			& & \textbf{AdaMergeX} & \textbf{\name} \\
			\midrule
			10 & 12000 & $0.05\pm0.001$ & $0.11\pm0.001$ \\
			50 & 12000 & $0.17\pm0.002$ & $0.34\pm0.002$ \\
			100 & 12000 & $0.35\pm0.001$ & $0.44\pm0.003$ \\
			500 & 12000 & $0.39\pm0.001$ & $0.49\pm0.002$ \\
			1000 & 12000 & $0.42\pm0.002$ & $0.52\pm0.001$ \\
			5000 & 12000 & $0.55\pm0.002$ & $0.64\pm0.003$ \\
			10000 & 12000 & $0.71\pm0.003$ & $0.79\pm0.002$ \\
			\midrule
			100 & 12000 & $0.35\pm0.001$ & $0.44\pm0.003$ \\
			100 & 5000 & $0.22\pm0.002$ & $0.41\pm0.002$ \\
			100 & 1000 &$0.11\pm0.028$ & $0.25\pm0.032$ \\
			100 & 500 & $0.04\pm0.025$ & $0.10\pm0.023$ \\
			\bottomrule
		\end{tabular}
	}
    \vspace*{-1mm}
	\tabcaption{NER performance comparison based on Macro-F1 between AdaMergeX and our approach (\name) with Hindi as high-resource and Marathi as low-resource language under varying dataset sizes.}
    \vspace*{-2mm}
	\label{tab:size_impact}
\end{table}

The second experiment, keeping LRL fixed at 100 instances while varying HRL size, reveals that AdaMergeX's performance degrades dramatically
with decreasing HRL data (from 0.35 F1 with 12,000 instances to just 0.04 F1
with 500 instances). While \name also shows decreased performance with less HRL
data, it maintains substantially better results (0.10 F1 even with just 500 HRL
instances) and demonstrates greater resilience to HRL data reduction. These
results highlight both \name's effectiveness at enabling cross-lingual
knowledge transfer and its superior ability to leverage limited high-resource
data compared to AdaMergeX. Our additional experiments on sentiment classification (details in Tables \ref{tab:sc_size_impact} and \ref{tab:ner_em_size_impact} in Appendix) reinforce these findings, with \name outperforming AdaMergeX by 14-28 percentage points for Hindi-Bangla and 12-27 percentage points for English-Bangla pairs across various dataset sizes.


We investigated whether increasing batch size during training improves training efficiency while maintaining performance. Training GETR-GAT with batch sizes of 60, 120, and 4096 on sentiment classification tasks across multiple language pairs, we found that smaller batches (120) achieved comparable performance to large batch sizes (4096) but required approximately 14\% more training steps to saturation. This is expected, as larger batches enable more direct token connections between low-resource and high-resource languages through the graph structure, allowing the model to learn global semantics more efficiently. Smaller batches require successive approximations through the 30\% random sampling strategy to establish these cross-lingual connections across epochs. These results demonstrate that GETR's effectiveness is robust to different batch sizes, though computational constraints typically limit practical batch sizes. Detailed results are presented in Sec.~\ref{section:ap_batch_size_effect} of Appendix.

To assess GETR's resilience to incomplete or noisy cross-lingual edge information, we conducted experiments by systematically reducing the percentage of cross-lingual token connections in the graph. Each typical batch contains 600--1000 token connections established through our translation mappings. We evaluated GETR-GAT performance with 100\%, 70\%, 50\%, 30\%, and 0\% of these connections, where 0\% reduces to standard Joint Training. Performance degraded gracefully with fewer connections, demonstrating that GETR maintains substantial gains even with 70\% of the edges (typically within 2--5 percentage points of full performance) and converges toward Joint Training baseline at 0\% edge retention. This robustness suggests that GETR does not require perfect bilingual lexicons and can function effectively even with partial or noisy translation dictionaries. Detailed results are presented in Sec.~\ref{section:ap_lexicon_effect} of Appendix.

\section{Conclusions}
\label{sec:conclusion}

In this paper, we addressed the challenge of cross-lingual knowledge transfer
for extreme low-resource scenarios.  We proposed two adopted baselines, HAL and TET, and a novel GETR mechanism. Experimental results demonstrate that while
traditional multilingual and cross-lingual models struggle with extreme data scarcity, our
proposed approaches effectively leverage knowledge from high-resource
languages.

Future work includes exploring linguistic insights and self-supervised pre-training strategies specific
to low-resource languages,
memory-optimized implementations of graph neural networks, and cross-lingual
transfer for a wider range of tasks and language pairs.

%

\section*{Acknowledgements} 
\label{sec:acknowledgement}

Following ARR's AI Writing/Coding Assistance Policy, we acknowledge using Claude 3.7 Sonnet \cite{anthropic2025claude37} for editorial assistance while maintaining full ownership of and responsibilities for all scientific contributions.

\section*{Limitations}
\label{sec:limitations}


While our proposed approaches demonstrate strong performance across different tasks and language pairs, we acknowledge certain aspects that present opportunities for future research. Our experiments primarily focus on Indian languages from both Indo-Aryan and Dravidian families, which could be extended to typologically more distant language pairs with different word orders or morphological systems in future work.

Although BhashaSetu is effective with minimal low-resource data (100 instances), we observe that transfer performance correlates with high-resource language data availability, a common pattern in transfer learning approaches. This relationship between source data volume and transfer effectiveness presents an interesting direction for developing more data-efficient transfer techniques.

The Token Embedding Transfer approach benefits from word-level translation capabilities between language pairs. While such resources exist for many language combinations, future work could explore unsupervised methods for establishing cross-lingual correspondences when traditional bilingual dictionaries are unavailable.

Our Graph-Enhanced Token Representation approach introduces additional computational complexity during training and inference due to graph construction operations and GNN computations compared to simpler methods. However, this computational investment delivers substantially improved performance (21-27 percentage points gain in F1 scores), representing a favorable trade-off in many practical scenarios. Future implementations could explore optimization techniques to reduce this overhead.

Finally, while we demonstrate effectiveness on classification tasks (sentiment analysis and NER), extending these approaches to generative tasks involving neural machine translation or summary generation represents a promising direction for future research. This would further validate the versatility of our framework across the broader NLP task spectrum.

\section*{Ethics Statement}
\label{sec:ethics}

This research aims to promote linguistic inclusivity by addressing the technological disparity between high-resource and low-resource languages. We acknowledge that NLP capabilities have predominantly benefited widely-spoken languages, potentially exacerbating digital divides along linguistic lines. All datasets used in our experiments are publicly available with appropriate citations, and we did not collect or annotate new data that might introduce privacy concerns. 

We recognize that transfer learning approaches may inadvertently propagate biases from source to target languages; however, our work takes a step toward mitigating representation disparities by enabling better performance with minimal labeled data in low-resource languages. Due to the focus on extremely low-resource settings (approximately 100 training instances), the computational requirements for target language adaptation were substantially lower than those typically needed for high-resource language model development, reducing the environmental impact compared to training large language models from scratch. While the GETR approaches do introduce additional computational overhead during the knowledge transfer process, the overall resource consumption remains modest relative to pre-training large multilingual models. This efficiency is particularly beneficial for researchers and practitioners with limited computational resources working on low-resource language technologies.

While we focused on Indian languages in this study, we believe that similar approaches could benefit other low-resource languages globally, contributing to more equitable language technology development. We emphasize that the performance improvements demonstrated should be considered within the context of the limitations described in our paper, and that practical applications would require careful consideration of cultural and linguistic nuances specific to each target community.

\bibliography{custom}

\appendix

\section*{Appendix}

\section{HAL}
\begin{figure}[t] 
	\centering 
	\includegraphics[width=0.99\linewidth]{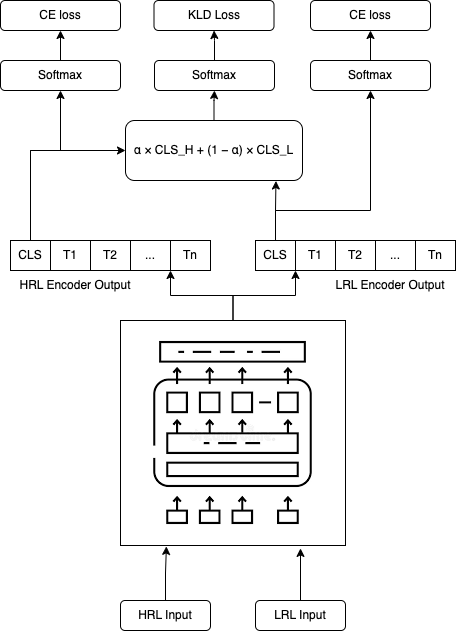} 
	\caption{Architecture incorporating the Hidden Augmentation Layer (HRL and LRL inputs are high- and low-resource language inputs respectively)} 
	\label{fig:ap_hidden_aug} 
\end{figure}

Figure~\ref{fig:ap_hidden_aug} illustrates our modified architecture incorporating the hidden augmentation layer. The framework can be further extended by adding multiple transformer layers above $E_M$ and performing augmentation at each layer's \texttt{CLS} output, thus enabling hierarchical knowledge fusion.

\section{Dynamic Mixing Coefficient Analysis}
\label{sec:appendix_dynamic_alpha}

We experimented with a dynamic mixing coefficient that changes linearly during model training, transitioning from 1 (favoring the high-resource language) to 0 (favoring the low-resource language). The definition of dynamic $\alpha$ is given by:

\[
\alpha = \frac{1 - s}{T}
\]

where $s$ represents the number of training steps and $T$ is the total number of training steps. This approach aims to gradually shift the model's focus from leveraging high-resource language knowledge during early training phases to increasingly relying on low-resource language-specific patterns as training progresses.

Implementing this dynamic mixing strategy, we found no major improvement over the existing HAL approach. Table~\ref{tab:dynamic_alpha_results} presents the comprehensive results for all high-resource and low-resource language pairs across both Sentiment Classification and Named Entity Recognition tasks, comparing the original HAL with fixed $\alpha=0.2$ to HAL with the dynamic $\alpha$ (HAL-DynamicAlpha).

\begin{table*}[t]
	\centering
	\small
	\resizebox{\linewidth}{!}{
		\begin{tabular}{llcccc}
			\toprule
			\multirow{2}{*}{\textbf{HRL}} & \multirow{2}{*}{\textbf{Method}} & \multicolumn{2}{c}{\textbf{Sentiment Classification}} & \multicolumn{2}{c}{\textbf{NER}} \\
			\cmidrule(lr){3-4} \cmidrule(lr){5-6}
			& & \textbf{Marathi as LRL} & \textbf{Bangla as LRL} & \textbf{Marathi as LRL} & \textbf{Malayalam as LRL} \\
			\midrule
			\multirow{2}{*}{English} & HAL & 0.60 $\pm$ 0.001 & 0.64 $\pm$ 0.001 & 0.32 $\pm$ 0.001 & 0.30 $\pm$ 0.003 \\
			& HAL-DynamicAlpha & 0.60 $\pm$ 0.002 & 0.64 $\pm$ 0.002 & 0.32 $\pm$ 0.001 & 0.30 $\pm$ 0.002 \\
			\midrule
			\multirow{2}{*}{Hindi} & HAL & 0.80 $\pm$ 0.005 & 0.72 $\pm$ 0.004 & 0.38 $\pm$ 0.001 & 0.32 $\pm$ 0.003 \\
			& HAL-DynamicAlpha & 0.80 $\pm$ 0.004 & 0.72 $\pm$ 0.003 & 0.38 $\pm$ 0.001 & 0.32 $\pm$ 0.003 \\
			\bottomrule
		\end{tabular}}
	\caption{Performance (Macro-F1 score) comparison of HAL and HAL-DynamicAlpha on Sentiment Classification and NER tasks across different high-resource and low-resource language pairs. Results are reported with mean and standard deviation based on independent runs.}
	\label{tab:dynamic_alpha_results}
\end{table*}

\subsection{Discussion}

The empirical results demonstrate that the dynamic mixing coefficient strategy yields negligible performance differences compared to the fixed $\alpha = 0.2$ configuration. Across all language pairs and both tasks, the performance metrics remain nearly identical between HAL and HAL-DynamicAlpha, with overlapping confidence intervals. This finding indicates that the computational overhead introduced by the dynamic schedule does not translate into meaningful improvements in model performance.

The stability of our results across the dynamic schedule further validates the robustness of the HAL approach with a fixed mixing coefficient of 0.2, as reported in the main paper. The dynamic strategy was motivated by the hypothesis that gradually shifting focus from high-resource to low-resource language knowledge during training might prevent negative transfer and improve generalization. However, our experiments suggest that a carefully tuned fixed mixing coefficient is sufficient to balance knowledge transfer from the high-resource language with preservation of low-resource language characteristics, making the added complexity of dynamic scheduling unnecessary. Note that while we implemented and evaluated a linear schedule for the dynamic mixing coefficient, more adaptive scheduling strategies such as uncertainty-based, performance-driven, or non-linear curricula were not explored in this work due to time and scope constraints. This clarifies the boundaries of our investigation for readers and future follow-up studies.

\subsection{Results on Sentiment Classification Task and NER}

We extensively evaluated our approach against multiple baselines, including parameter-efficient fine-tuning methods and XLM-R variants. For XLM-R, we tested: (1) XLMFT-LRL: fine-tuning only on low-resource data; (2) XLMFT-HRLRL: joint fine-tuning on both language datasets; (3) XLMFT-HRL2LRL: sequential fine-tuning on high-resource followed by low-resource data; and (4) XLMFT-HRLAdapLRL: fine-tuning on high-resource data followed by adapter-based fine-tuning on low-resource data with frozen base weights. Additionally, we evaluated LoRA and AdaMergeX as parameter-efficient alternatives, and HAL-LRL which applies augmentation only within the low-resource language. Our results show XLM-R variants perform comparably to Joint Training across all configurations, while HAL-LRL shows no improvement over Joint Training due to limited augmentation diversity in the extremely small low-resource dataset.

\begin{table*}[t]
	\centering
	\small
	\resizebox{\linewidth}{!}{
		\begin{tabular}{llcccc}
			\toprule
			\multirow{2}{*}{\textbf{HRL}} & \multirow{2}{*}{\textbf{Method}} & \multicolumn{2}{c}{\textbf{Sentiment Classification}} & \multicolumn{2}{c}{\textbf{NER}} \\
			\cmidrule(lr){3-4} \cmidrule(lr){5-6}
			& & \textbf{Marathi as LRL} & \textbf{Bangla as LRL} & \textbf{Marathi as LRL} & \textbf{Malayalam as LRL} \\
			\midrule
			- & Scratch Training & 0.33 $\pm$ 0.000 & 0.33 $\pm$ 0.000 & 0.03 $\pm$ 0.073 & 0.03 $\pm$ 0.073 \\
			\midrule
			\multirow{20}{*}{English} & Joint Training & 0.53 $\pm$ 0.002 & 0.63 $\pm$ 0.001 & 0.29 $\pm$ 0.001 & 0.26 $\pm$ 0.002 \\
			& Adapter Finetuning & 0.53 $\pm$ 0.001 & 0.63 $\pm$ 0.002 & 0.29 $\pm$ 0.001 & 0.26 $\pm$ 0.001 \\
			& XLMFT-LRL & 0.49 $\pm$ 0.002 & 0.60 $\pm$ 0.004 & 0.27 $\pm$ 0.005 & 0.23 $\pm$ 0.003 \\
			& XLMFT-HRLRL & 0.53 $\pm$ 0.002 & 0.63 $\pm$ 0.001 & 0.29 $\pm$ 0.001 & 0.26 $\pm$ 0.001 \\
			& XLMFT-HRL2LRL & 0.52 $\pm$ 0.002 & 0.63 $\pm$ 0.004 & 0.28 $\pm$ 0.003 & 0.24 $\pm$ 0.003 \\
			& XLMFT-HRLAdapLRL & 0.53 $\pm$ 0.001 & 0.63 $\pm$ 0.002 & 0.29 $\pm$ 0.002 & 0.26 $\pm$ 0.001 \\
			& LoRA & 0.54 $\pm$ 0.001 & 0.63 $\pm$ 0.001 & 0.29 $\pm$ 0.001 & 0.27 $\pm$ 0.001 \\
			& AdaMergeX & 0.55 $\pm$ 0.001 & 0.63 $\pm$ 0.001 & 0.29 $\pm$ 0.001 & 0.28 $\pm$ 0.002 \\
			& HAL-LRL & 0.52 $\pm$ 0.002 & 0.63 $\pm$ 0.001 & 0.29 $\pm$ 0.001 & 0.26 $\pm$ 0.001 \\
			& HAL & 0.60 $\pm$ 0.001 & 0.64 $\pm$ 0.001 & 0.32 $\pm$ 0.001 & 0.30 $\pm$ 0.003 \\
			& HAL + TET & 0.63 $\pm$ 0.001 & 0.65 $\pm$ 0.003 & 0.33 $\pm$ 0.001 & 0.31 $\pm$ 0.002 \\
			& GETR-GCN & 0.69 $\pm$ 0.002 & 0.68 $\pm$ 0.001 & 0.36 $\pm$ 0.001 & 0.37 $\pm$ 0.001 \\
			& GETR-GCN + TET & 0.68 $\pm$ 0.001 & 0.69 $\pm$ 0.003 & 0.36 $\pm$ 0.001 & 0.37 $\pm$ 0.002 \\
			& GETR-GCN + HAL & 0.70 $\pm$ 0.001 & 0.70 $\pm$ 0.002 & 0.39 $\pm$ 0.001 & 0.43 $\pm$ 0.003 \\
			& GETR-GCN + HAL + TET & 0.70 $\pm$ 0.002 & 0.70 $\pm$ 0.001 & 0.39 $\pm$ 0.001 & 0.43 $\pm$ 0.002 \\
			& GETR-GAT & 0.73 $\pm$ 0.001 & 0.72 $\pm$ 0.001 & 0.40 $\pm$ 0.001 & 0.46 $\pm$ 0.001 \\
			& GETR-GAT + TET & 0.74 $\pm$ 0.001 & 0.73 $\pm$ 0.002 & 0.40 $\pm$ 0.001 & 0.47 $\pm$ 0.003 \\
			& GETR-GAT + HAL & \textbf{0.75 $\pm$ 0.001} & 0.74 $\pm$ 0.001 & \textbf{0.40 $\pm$ 0.001} & 0.51 $\pm$ 0.002 \\
			& GETR-GAT + HAL + TET & 0.74 $\pm$ 0.001 & \textbf{0.75 $\pm$ 0.001} & 0.40 $\pm$ 0.001 & \textbf{0.52 $\pm$ 0.001} \\
			\midrule
			\multirow{20}{*}{Hindi} & Joint Training & 0.75 $\pm$ 0.004 & 0.67 $\pm$ 0.003 & 0.35 $\pm$ 0.002 & 0.28 $\pm$ 0.002 \\
			& Adapter Finetuning & 0.74 $\pm$ 0.002 & 0.67 $\pm$ 0.002 & 0.34 $\pm$ 0.003 & 0.28 $\pm$ 0.002 \\
			& XLMFT-LRL & 0.71 $\pm$ 0.003 & 0.62 $\pm$ 0.005 & 0.30 $\pm$ 0.004 & 0.26 $\pm$ 0.004 \\
			& XLMFT-HRLRL & 0.75 $\pm$ 0.001 & 0.67 $\pm$ 0.001 & 0.34 $\pm$ 0.001 & 0.28 $\pm$ 0.001 \\
			& XLMFT-HRL2LRL & 0.75 $\pm$ 0.003 & 0.67 $\pm$ 0.004 & 0.34 $\pm$ 0.004 & 0.27 $\pm$ 0.001 \\
			& XLMFT-HRLAdapLRL & 0.74 $\pm$ 0.002 & 0.67 $\pm$ 0.002 & 0.34 $\pm$ 0.003 & 0.28 $\pm$ 0.002 \\
			& LoRA & 0.75 $\pm$ 0.001 & 0.68 $\pm$ 0.001 & 0.35 $\pm$ 0.002 & 0.28 $\pm$ 0.001 \\
			& AdaMergeX & 0.76 $\pm$ 0.001 & 0.69 $\pm$ 0.001 & 0.36 $\pm$ 0.001 & 0.28 $\pm$ 0.001 \\
			& HAL-LRL & 0.75 $\pm$ 0.003 & 0.67 $\pm$ 0.002 & 0.35 $\pm$ 0.001 & 0.27 $\pm$ 0.001 \\
			& HAL & 0.80 $\pm$ 0.005 & 0.72 $\pm$ 0.004 & 0.38 $\pm$ 0.001 & 0.32 $\pm$ 0.003 \\
			& HAL + TET & - & 0.73 $\pm$ 0.002 & - & 0.32 $\pm$ 0.002 \\
			& GETR-GCN & 0.82 $\pm$ 0.001 & 0.75 $\pm$ 0.001 & 0.42 $\pm$ 0.002 & 0.38 $\pm$ 0.001 \\
			& GETR-GCN + TET & - & 0.75 $\pm$ 0.002 & - & 0.38 $\pm$ 0.002 \\
			& GETR-GCN + HAL & 0.83 $\pm$ 0.002 & 0.76 $\pm$ 0.001 & 0.43 $\pm$ 0.001 & 0.44 $\pm$ 0.003 \\
			& GETR-GCN + HAL + TET & - & 0.76 $\pm$ 0.002 & - & 0.44 $\pm$ 0.002 \\
			& GETR-GAT & 0.85 $\pm$ 0.001 & 0.79 $\pm$ 0.002 & 0.44 $\pm$ 0.001 & 0.48 $\pm$ 0.001 \\
			& GETR-GAT + TET & - & 0.80 $\pm$ 0.001 & - & 0.49 $\pm$ 0.003 \\
			& GETR-GAT + HAL & \textbf{0.87 $\pm$ 0.001} & 0.80 $\pm$ 0.003 & \textbf{0.44 $\pm$ 0.001} & 0.53 $\pm$ 0.002 \\
			& GETR-GAT + HAL + TET & - & \textbf{0.81 $\pm$ 0.002} & - & \textbf{0.55 $\pm$ 0.001} \\
			\bottomrule
	\end{tabular}}
	\caption{Performance (Macro-F1 score) comparison of different training approaches on sentiment classification and NER datasets when Hindi and English are considered as HRL and Marathi, Bangla and Malayalam as LRL. The mean and standard deviation numbers are reported based on 5 independent runs.}
	\label{ap:all_results}
\end{table*}

To understand the impact of mixing coefficient $\alpha$ in Hidden Augmentation Layer (HAL), we conducted experiments with different $\alpha$ values ranging from 0.1 to 0.8 (Table \ref{tab:hal_alpha_results}). For both English and Hindi as high-resource languages, $\alpha$=0.2 yields the best performance, achieving accuracy/F1 scores of 0.610/0.590 and 0.860/0.860 respectively. The performance gradually degrades as $\alpha$ increases, with a more pronounced decline after $\alpha$=0.5. This suggests that while knowledge from the high-resource language provides useful linguistic patterns and Sentiment structures, excessive reliance on it diminishes the model's ability to capture the unique characteristics and nuances of the low-resource language. The optimal performance at $\alpha$=0.2 indicates that a balanced approach, where the model primarily learns from the low-resource language while leveraging complementary features from the high-resource language, is most effective. Notably, even with declining performance at higher $\alpha$ values, the model maintains reasonable performance (minimum accuracy of 0.590 for English and 0.830 for Hindi as HRL), indicating the robustness of the HAL approach across different mixing ratios.

\begin{table}[htbp]
	\footnotesize
	\centering
	\resizebox{\linewidth}{!}{
		\begin{tabular}{l|lrrr}
			\toprule
			\multirow{2}{*}{\textbf{HRL}} & \multirow{2}{*}{\textbf{LRL}} & \multirow{2}{*}{\textbf{$\alpha$}} & \multicolumn{2}{c}{\textbf{Metrics}} \\
			\cline{4-5}
			&  &  & \textbf{Accuracy} & \textbf{F1} \\
			\toprule
			\multirow{8}{*}{English} & \multirow{8}{*}{Marathi} & 0.1 & $0.602{\pm}0.004$ & $0.582{\pm}0.005$ \\
			& & \textbf{0.2} & $\mathbf{0.610{\pm}0.004}$ & $\mathbf{0.590{\pm}0.005}$ \\
			& & 0.3 & $0.605{\pm}0.003$ & $0.578{\pm}0.004$ \\
			& & 0.4 & $0.598{\pm}0.004$ & $0.571{\pm}0.005$ \\
			& & 0.5 & $0.595{\pm}0.005$ & $0.565{\pm}0.004$ \\
			& & 0.6 & $0.592{\pm}0.004$ & $0.558{\pm}0.005$ \\
			& & 0.7 & $0.591{\pm}0.005$ & $0.552{\pm}0.004$ \\
			& & 0.8 & $0.590{\pm}0.004$ & $0.550{\pm}0.005$ \\
			\midrule
			\multirow{8}{*}{Hindi} & \multirow{8}{*}{Marathi} & 0.1 & $0.852{\pm}0.004$ & $0.848{\pm}0.005$ \\
			& & \textbf{0.2} & $\mathbf{0.860{\pm}0.003}$ & $\mathbf{0.860{\pm}0.005}$ \\
			& & 0.3 & $0.848{\pm}0.004$ & $0.845{\pm}0.004$ \\
			& & 0.4 & $0.842{\pm}0.005$ & $0.840{\pm}0.005$ \\
			& & 0.5 & $0.838{\pm}0.004$ & $0.835{\pm}0.004$ \\
			& & 0.6 & $0.834{\pm}0.005$ & $0.832{\pm}0.005$ \\
			& & 0.7 & $0.832{\pm}0.004$ & $0.831{\pm}0.004$ \\
			& & 0.8 & $0.830{\pm}0.005$ & $0.830{\pm}0.005$ \\
			\bottomrule
	\end{tabular}}
	\caption{Performance comparison of HAL approach with different high-resource languages and varying $\alpha$ values. HRL: High Resource Language, LRL: Low Resource Language}
	\label{tab:hal_alpha_results}
\end{table}

We analyzed the impact of HAL depth by varying the number of layers from 1 to 6 (Table \ref{tab:hal_depth_results}). For both English and Hindi as high-resource languages, 2 HAL layers yield optimal performance (accuracy/F1: 0.610/0.590 and 0.860/0.860 respectively), with secondary peaks at depth 4 for English (0.598/0.582) and depth 5 for Hindi (0.848/0.845), suggesting that while multiple HAL layers aid in knowledge transfer, excessive depth might lead to over-abstraction of features. Similarly, for both GETR-GCN and GETR-GAT approaches, three GNN layers demonstrated the best performance on the test set metrics, indicating an optimal depth for graph-based token interaction.

\begin{table}[b]
	\centering
	\resizebox{\linewidth}{!}{
		\begin{tabular}{l|lrrr}
			\toprule
			\multirow{2}{*}{\textbf{HRL}} & \multirow{2}{*}{\textbf{LRL}} & \textbf{HAL} & \multicolumn{2}{c}{\textbf{Metrics}} \\
			\cline{4-5}
			&  & \textbf{Depth} & \textbf{Accuracy} & \textbf{F1} \\
			\toprule
			\multirow{6}{*}{English} & \multirow{6}{*}{Marathi} & 1 & $0.592{\pm}0.004$ & $0.575{\pm}0.005$ \\
			& & \textbf{2} & $\mathbf{0.610{\pm}0.004}$ & $\mathbf{0.590{\pm}0.005}$ \\
			& & 3 & $0.588{\pm}0.003$ & $0.562{\pm}0.004$ \\
			& & 4 & $0.598{\pm}0.004$ & $0.582{\pm}0.005$ \\
			& & 5 & $0.575{\pm}0.005$ & $0.545{\pm}0.004$ \\
			& & 6 & $0.570{\pm}0.004$ & $0.540{\pm}0.005$ \\
			\midrule
			\multirow{6}{*}{Hindi} & \multirow{6}{*}{Marathi} & 1 & $0.842{\pm}0.004$ & $0.838{\pm}0.005$ \\
			& & \textbf{2} & $\mathbf{0.860{\pm}0.003}$ & $\mathbf{0.860{\pm}0.005}$ \\
			& & 3 & $0.835{\pm}0.004$ & $0.832{\pm}0.004$ \\
			& & 4 & $0.825{\pm}0.005$ & $0.818{\pm}0.005$ \\
			& & 5 & $0.848{\pm}0.004$ & $0.845{\pm}0.004$ \\
			& & 6 & $0.810{\pm}0.005$ & $0.800{\pm}0.005$ \\
			\bottomrule
	\end{tabular}}
	\caption{Impact of HAL depth on model performance. HRL: High Resource Language, LRL: Low Resource Language}
	\label{tab:hal_depth_results}
\end{table}

We extended our robustness evaluation to sentiment classification with Bangla as the low-resource language, testing both Hindi and English as high-resource languages (Table~\ref{tab:sc_size_impact}). The results reveal consistent advantages for \name across all data configurations. With minimal low-resource data (10 instances), Joint Training achieves only 0.33 macro-F1 for both HRLs, while \name reaches 0.61 with Hindi and 0.60 with English—an approximately 85\% improvement. This advantage persists across all LRL sizes, though the gap narrows as training data increases. Hindi consistently outperforms English as the high-resource language, with \name reaching 0.94 F1 using Hindi versus 0.89 F1 using English at 8,000 LRL instances.

The fixed LRL experiments (100 instances) with varying HRL size reveal \name's remarkable resilience to limited high-resource data. With just 500 HRL instances, \name maintains 0.62 F1 (Hindi) and 0.57 F1 (English), while Joint Training drops to 0.43 and 0.41 respectively. Most impressively, \name with just 1,000 Hindi instances (0.73 F1) outperforms Joint Training with the full 12,000 instances (0.67 F1). These results demonstrate \name's exceptional data efficiency in leveraging limited resources for cross-lingual transfer and confirm its effectiveness across both NER and sentiment classification tasks, regardless of the specific high-resource language used.

\begin{table}[tb]
	\centering
	\resizebox{\columnwidth}{!}{
		\begin{tabular}{lrrcc}
			\toprule
			\textbf{HRL} & \textbf{HRL Size} & \textbf{LRL Size} & \textbf{\shortstack{Macro F1 + JT}} & \textbf{\shortstack{Macro F1 + \name}} \\
			\toprule
			\multicolumn{5}{c}{\textit{Fixed HRL Size, Varying LRL Size}} \\
			\midrule
			Hindi & 12000 & 10 & $0.33\pm0.001$ & $0.61\pm0.001$ \\
			Hindi & 12000 & 50 & $0.51\pm0.002$ & $0.72\pm0.002$ \\
			Hindi & 12000 & 100 & $0.67\pm0.001$ & $0.81\pm0.003$ \\
			Hindi & 12000 & 500 & $0.69\pm0.001$ & $0.83\pm0.002$ \\
			Hindi & 12000 & 1000 & $0.73\pm0.002$ & $0.87\pm0.001$ \\
			Hindi & 12000 & 5000 & $0.79\pm0.002$ & $0.92\pm0.003$ \\
			Hindi & 12000 & 8000 & $0.82\pm0.003$ & $0.94\pm0.002$ \\
			\midrule
			English & 12000 & 10 & $0.33\pm0.001$ & $0.60\pm0.001$ \\
			English & 12000 & 50 & $0.49\pm0.002$ & $0.68\pm0.002$ \\
			English & 12000 & 100 & $0.63\pm0.001$ & $0.75\pm0.003$ \\
			English & 12000 & 500 & $0.65\pm0.001$ & $0.78\pm0.002$ \\
			English & 12000 & 1000 & $0.69\pm0.002$ & $0.81\pm0.001$ \\
			English & 12000 & 5000 & $0.74\pm0.002$ & $0.87\pm0.003$ \\
			English & 12000 & 8000 & $0.78\pm0.003$ & $0.89\pm0.002$ \\
			\midrule
			\multicolumn{5}{c}{\textit{Fixed LRL Size, Varying HRL Size}} \\
			\midrule
			Hindi & 12000 & 100 & $0.67\pm0.001$ & $0.81\pm0.003$ \\
			Hindi & 5000 & 100 & $0.61\pm0.002$ & $0.76\pm0.002$ \\
			Hindi & 1000 & 100 & $0.52\pm0.023$ & $0.73\pm0.003$ \\
			Hindi & 500 & 100 & $0.43\pm0.022$ & $0.62\pm0.006$ \\
			\midrule
			English & 12000 & 100 & $0.63\pm0.001$ & $0.75\pm0.003$ \\
			English & 5000 & 100 & $0.55\pm0.002$ & $0.71\pm0.002$ \\
			English & 1000 & 100 & $0.50\pm0.023$ & $0.65\pm0.003$ \\
			English & 500 & 100 & $0.41\pm0.022$ & $0.57\pm0.006$ \\
			\bottomrule
		\end{tabular}
	}
	\caption{Sentiment Classification performance comparison based on Macro-F1 between Joint Training (JT) and our approach (\name) with Hindi and English as high-resource and Bangla as low-resource language under varying dataset sizes.}
	\label{tab:sc_size_impact}
\end{table}

To evaluate the robustness of our approach on sentiment classification, we conducted extensive experiments varying dataset sizes with both Hindi and English as high-resource languages for Bangla (Table~\ref{tab:sc_size_impact}). With Hindi as HRL, \name demonstrates remarkable effectiveness, achieving 0.61 macro-F1 with just 10 LRL instances compared to Joint Training's 0.33—an improvement of 28 percentage points. This advantage persists as LRL size increases, maintaining improvements of 12-21 percentage points up to 8,000 instances (the maximum available in our Bangla dataset), where \name achieves 0.94 macro-F1 compared to Joint Training's 0.82.

Similar patterns emerge with English as HRL, though with slightly lower absolute performance due to script differences. \name achieves 0.60 macro-F1 with 10 LRL instances (27 percentage points over Joint Training) and maintains substantial improvements through 8,000 instances (0.89 vs 0.78 macro-F1). The fixed LRL experiments (100 instances) reveal \name's superior resilience to HRL data reduction: with Hindi, performance drops from 0.81 to 0.62 macro-F1 as HRL size decreases from 12,000 to 500, while Joint Training falls more sharply from 0.67 to 0.43. English shows similar trends, with \name maintaining better performance (0.75 to 0.57) compared to Joint Training's steeper decline (0.63 to 0.41). These results demonstrate \name's effectiveness across different data regimes and language pairs, with particularly strong performance when languages share scripts.

\begin{table}[htbp]
	\centering
	\resizebox{\columnwidth}{!}{
		\begin{tabular}{rrcc}
			\toprule
			\textbf{LRL Size} & \textbf{HRL Size} & \textbf{\shortstack{Macro F1 + AdaMergeX}} & \textbf{\shortstack{Macro F1 + \name}} \\
			\toprule
			\multicolumn{4}{c}{\textit{Fixed HRL Size, Varying LRL Size}} \\
			\midrule
			10 & 12000 & $0.02\pm0.001$ & $0.11\pm0.001$ \\
			50 & 12000 & $0.13\pm0.002$ & $0.34\pm0.002$ \\
			100 & 12000 & $0.29\pm0.001$ & $0.40\pm0.003$ \\
			500 & 12000 & $0.34\pm0.001$ & $0.46\pm0.002$ \\
			1000 & 12000 & $0.39\pm0.002$ & $0.49\pm0.001$ \\
			5000 & 12000 & $0.51\pm0.002$ & $0.57\pm0.002$ \\
			10000 & 12000 & $0.64\pm0.001$ & $0.73\pm0.001$ \\
			\midrule
			\multicolumn{4}{c}{\textit{Fixed LRL Size, Varying HRL Size}} \\
			\midrule
			100 & 12000 & $0.29\pm0.001$ & $0.40\pm0.003$ \\
			100 & 5000 & $0.18\pm0.002$ & $0.34\pm0.002$ \\
			100 & 1000 &$0.07\pm0.025$ & $0.20\pm0.034$ \\
			100 & 500 & $0.03\pm0.022$ & $0.07\pm0.031$ \\
			\bottomrule
		\end{tabular}
	}
	\caption{NER performance comparison based on Macro-F1 between AdaMergeX and our approach (\name) with English as high-resource and Marathi as low-resource language under varying dataset sizes.}
	\label{tab:ner_em_size_impact}
\end{table}

\section{Detailed Ablation Studies}

\subsection{Effect of Batch Size on Training Efficiency}
\label{section:ap_batch_size_effect}

To understand the relationship between batch size and training efficiency, we trained GETR-GAT models using batch sizes of 60, 120, and 4096 on a \texttt{p5en.48xlarge} instance. Each batch of size \textit{B} contained all 100 low-resource language instances available in the epoch, with the remaining \textit{B}-100 instances sampled from high-resource languages. Of these high-resource samples, 70\% were selected based on maximum token overlap with low-resource instances, while the remaining 30\% were randomly selected to ensure gradual establishment of cross-lingual connections across epochs.

The motivation for larger batch sizes stems from GETR's graph-based token connection mechanism. With batch size 4096, nearly all unique tokens from low-resource languages can be connected to high-resource tokens simultaneously, allowing the model to learn global cross-lingual semantics directly. In contrast, smaller batches (e.g., 120) establish these connections incrementally through successive epochs via the 30\% random sampling strategy, requiring more steps to achieve saturation.

Our experiments across sentiment classification and NER tasks with Hindi and English as high-resource languages and Marathi and Malayalam as low-resource languages reveal consistent patterns. Training with batch size 4096 achieved comparable performance to batch size 120 while requiring approximately 14\% fewer training steps to reach convergence. Batch size 60 required the most steps but maintained similar final performance, confirming that GETR's effectiveness is robust across different batch sizes. However, due to computational constraints, most practitioners will be limited to smaller batch sizes. Tables~\ref{tab:batch_size_sc_marathi}, \ref{tab:batch_size_sc_bangla}, \ref{tab:batch_size_ner_marathi} and \ref{tab:batch_size_ner_malayalam} present detailed results across all tasks and language pairs.

\begin{table}[t]
	\centering
	\small
	\resizebox{\linewidth}{!}{
		\begin{tabular}{llrrc}
			\toprule
			\multirow{2}{*}{\textbf{HRL}} & \multirow{2}{*}{\textbf{Method}} & \multicolumn{1}{c}{\textbf{Batch}} & \multicolumn{1}{c}{\textbf{Training}} & \textbf{Sentiment Classification} \\
			\cmidrule(lr){5-5}
			& & \multicolumn{1}{c}{\textbf{Size}} & \multicolumn{1}{c}{\textbf{Steps}} & \textbf{Marathi as LRL} \\
			\midrule
			\multirow{3}{*}{English} & GETR-GAT & 60 & 7141 & $0.73 \pm 0.001$ \\
			& GETR-GAT & 120 & 5416 & $0.73 \pm 0.001$ \\
			& GETR-GAT & 4096 & 4610 & $0.73 \pm 0.002$ \\
			\midrule
			\multirow{3}{*}{Hindi} & GETR-GAT & 60 & 7002 & $0.85 \pm 0.001$ \\
			& GETR-GAT & 120 & 5319 & $0.85 \pm 0.001$ \\
			& GETR-GAT & 4096 & 4220 & $0.85 \pm 0.003$ \\
			\bottomrule
		\end{tabular}
	}
	\caption{GETR-GAT performance (Macro-F1 score) on Sentiment Classification with Marathi as low-resource language across different batch sizes.}
	\label{tab:batch_size_sc_marathi}
\end{table}

\begin{table}[t]
	\centering
	\small
	\resizebox{\linewidth}{!}{
		\begin{tabular}{llrrc}
			\toprule
			\multirow{2}{*}{\textbf{HRL}} & \multirow{2}{*}{\textbf{Method}} & \multicolumn{1}{c}{\textbf{Batch}} & \multicolumn{1}{c}{\textbf{Training}} & \textbf{Sentiment Classification} \\
			\cmidrule(lr){5-5}
			& & \multicolumn{1}{c}{\textbf{Size}} & \multicolumn{1}{c}{\textbf{Steps}} & \textbf{Bangla as LRL} \\
			\midrule
			\multirow{3}{*}{English} & GETR-GAT & 60 & 6981 & $0.72 \pm 0.001$ \\
			& GETR-GAT & 120 & 5501 & $0.72 \pm 0.001$ \\
			& GETR-GAT & 4096 & 4690 & $0.72 \pm 0.002$ \\
			\midrule
			\multirow{3}{*}{Hindi} & GETR-GAT & 60 & 6993 & $0.79 \pm 0.001$ \\
			& GETR-GAT & 120 & 5561 & $0.79 \pm 0.002$ \\
			& GETR-GAT & 4096 & 4399 & $0.79 \pm 0.002$ \\
			\bottomrule
		\end{tabular}
	}
	\caption{GETR-GAT performance (Macro-F1 score) on Sentiment Classification with Bangla as low-resource language across different batch sizes.}
	\label{tab:batch_size_sc_bangla}
\end{table}

\begin{table}[t]
	\centering
	\small
	\resizebox{\linewidth}{!}{
		\begin{tabular}{llrrc}
			\toprule
			\multirow{2}{*}{\textbf{HRL}} & \multirow{2}{*}{\textbf{Method}} & \multicolumn{1}{c}{\textbf{Batch}} & \multicolumn{1}{c}{\textbf{Training}} & \textbf{NER} \\
			\cmidrule(lr){5-5}
			& & \multicolumn{1}{c}{\textbf{Size}} & \multicolumn{1}{c}{\textbf{Steps}} & \textbf{Marathi as LRL} \\
			\midrule
			\multirow{3}{*}{English} & GETR-GAT & 60 & 6981 & $0.40 \pm 0.001$ \\
			& GETR-GAT & 120 & 5501 & $0.40 \pm 0.001$ \\
			& GETR-GAT & 4096 & 4690 & $0.40 \pm 0.002$ \\
			\midrule
			\multirow{3}{*}{Hindi} & GETR-GAT & 60 & 6993 & $0.44 \pm 0.001$ \\
			& GETR-GAT & 120 & 5561 & $0.44 \pm 0.001$ \\
			& GETR-GAT & 4096 & 4399 & $0.44 \pm 0.001$ \\
			\bottomrule
		\end{tabular}
	}
	\caption{GETR-GAT performance (Macro-F1 score) on NER with Marathi as low-resource language across different batch sizes.}
	\label{tab:batch_size_ner_marathi}
\end{table}

\begin{table}[t]
	\centering
	\small
	\resizebox{\linewidth}{!}{
		\begin{tabular}{llrrc}
			\toprule
			\multirow{2}{*}{\textbf{HRL}} & \multirow{2}{*}{\textbf{Method}} & \multicolumn{1}{c}{\textbf{Batch}} & \multicolumn{1}{c}{\textbf{Training}} & \textbf{NER} \\
			\cmidrule(lr){5-5}
			& & \multicolumn{1}{c}{\textbf{Size}} & \multicolumn{1}{c}{\textbf{Steps}} & \textbf{Malayalam as LRL} \\
			\midrule
			\multirow{3}{*}{English} & GETR-GAT & 60 & 6882 & $0.46 \pm 0.001$ \\
			& GETR-GAT & 120 & 5483 & $0.46 \pm 0.001$ \\
			& GETR-GAT & 4096 & 4552 & $0.46 \pm 0.002$ \\
			\midrule
			\multirow{3}{*}{Hindi} & GETR-GAT & 60 & 7114 & $0.48 \pm 0.001$ \\
			& GETR-GAT & 120 & 5632 & $0.48 \pm 0.001$ \\
			& GETR-GAT & 4096 & 4781 & $0.48 \pm 0.003$ \\
			\bottomrule
		\end{tabular}
	}
	\caption{GETR-GAT performance (Macro-F1 score) on NER with Malayalam as low-resource language across different batch sizes.}
	\label{tab:batch_size_ner_malayalam}
\end{table}

\subsection{Robustness to Incomplete Bilingual Lexicons}
\label{section:ap_lexicon_effect}

To evaluate GETR's resilience to incomplete or noisy cross-lingual edge information, we systematically ablated the percentage of cross-lingual token connections in the graph. In our standard implementation, each batch typically contains 600--1000 token connections established through our translation mappings (using pymultidictionary and manual verification). We conducted experiments retaining 100\%, 70\%, 50\%, 30\%, and 0\% of these connections through random removal.

The 0\% condition serves as a critical baseline: when all cross-lingual edges are removed, GETR-GAT reduces to standard Joint Training, as tokens are only connected within sentences (not across languages). This allows us to quantify the precise contribution of cross-lingual knowledge transfer to overall performance.

The results demonstrate that GETR's performance degrades gracefully as fewer edges are retained. With 70\% edge retention, performance typically drops by only 2--5 percentage points, indicating substantial robustness to incomplete lexicons. Even at 50\% retention, models maintain meaningful improvements over Joint Training baselines. The gradual degradation pattern confirms that the approach does not rely on perfect bilingual lexicons and can function effectively with partial translation information. At 0\% edge retention, performance converges precisely to the Joint Training baseline across all tasks and language pairs, validating our hypothesis that cross-lingual edges are the key mechanism enabling GETR's knowledge transfer capabilities. Table~\ref{tab:lexicon_robustness} presents comprehensive results across sentiment classification and NER tasks.

\begin{table*}[!htbp]
	\centering
	\small
		\begin{tabular}{llrcccccc}
			\toprule
			\multirow{2}{*}{\textbf{HRL}} & \multirow{2}{*}{\textbf{Method}} & \multirow{2}{*}{\textbf{Edge \%}}  & \multicolumn{2}{c}{\textbf{Sentiment Classification}} & \multicolumn{2}{c}{\textbf{NER}} \\
			\cmidrule(lr){4-5} \cmidrule(lr){6-7}
			& & & \textbf{Marathi} & \textbf{Bangla} & \textbf{Marathi} & \textbf{Malayalam} \\
			\midrule
			\multirow{6}{*}{English} & Joint Training & - &  $0.53 \pm 0.002$ & $0.63 \pm 0.001$ & $0.29 \pm 0.001$ & $0.26 \pm 0.002$ \\
			& AdaMergeX & - &  $0.55 \pm 0.001$ & $0.63 \pm 0.001$ & $0.29 \pm 0.001$ & $0.28 \pm 0.002$ \\
			& GETR-GAT & 100\% & $0.73 \pm 0.001$ & $0.72 \pm 0.001$ & $0.40 \pm 0.001$ & $0.46 \pm 0.001$ \\
			& GETR-GAT & 70\%  & $0.67 \pm 0.002$ & $0.68 \pm 0.001$ & $0.38 \pm 0.001$ & $0.41 \pm 0.001$ \\
			& GETR-GAT & 50\%  & $0.64 \pm 0.001$ & $0.66 \pm 0.003$ & $0.35 \pm 0.003$ & $0.38 \pm 0.001$ \\
			& GETR-GAT & 30\%  & $0.58 \pm 0.002$ & $0.63 \pm 0.001$ & $0.31 \pm 0.001$ & $0.32 \pm 0.002$ \\
			& GETR-GAT & 0\%  & $0.55 \pm 0.003$ & $0.62 \pm 0.001$ & $0.28 \pm 0.002$ & $0.28 \pm 0.002$ \\
			\midrule
			\multirow{6}{*}{Hindi} & Joint Training & - & $0.75 \pm 0.004$ & $0.67 \pm 0.003$ & $0.35 \pm 0.002$ & $0.28 \pm 0.002$ \\
			& AdaMergeX & - & $0.76 \pm 0.001$ & $0.69 \pm 0.001$ & $0.30 \pm 0.001$ & $0.28 \pm 0.001$ \\
			& GETR-GAT & 100\%  & $0.85 \pm 0.001$ & $0.79 \pm 0.002$ & $0.44 \pm 0.001$ & $0.48 \pm 0.001$ \\
			& GETR-GAT & 70\%  & $0.83 \pm 0.002$ & $0.76 \pm 0.002$ & $0.40 \pm 0.001$ & $0.43 \pm 0.001$ \\
			& GETR-GAT & 50\%  & $0.80 \pm 0.001$ & $0.72 \pm 0.003$ & $0.36 \pm 0.002$ & $0.35 \pm 0.001$ \\
			& GETR-GAT & 30\%  & $0.77 \pm 0.002$ & $0.71 \pm 0.002$ & $0.31 \pm 0.001$ & $0.32 \pm 0.003$ \\
			& GETR-GAT & 0\%  & $0.75 \pm 0.001$ & $0.69 \pm 0.001$ & $0.30 \pm 0.002$ & $0.29 \pm 0.001$ \\
			\bottomrule
		\end{tabular}
	\caption{Robustness of GETR-GAT to incomplete bilingual lexicons (Macro-F1 score). Performance is measured when retaining 100\%, 70\%, 50\%, 30\%, and 0\% of cross-lingual token connections. Sentiment Classification spans Marathi and Bangla low-resource languages, while NER spans Marathi and Malayalam. Average token connections per batch are shown for each edge retention percentage.}
	\label{tab:lexicon_robustness}
\end{table*}

\subsection{Computational Cost and Environmental Impact}

GETR-GAT incurs modest computational overhead compared to Joint Training while delivering substantial performance gains. Training on AWS p4de.24xlarge instances requires approximately 11\% additional time per epoch ($\approx$ 50 minutes vs. $\approx$ 45 minutes for Joint Training) and 8\% more peak GPU memory ($\approx$ 41 GB vs. $\approx$ 38 GB) due to graph neural network computations and neighborhood construction. Over 50 training epochs, the total training time increases from 37.5 hours to 41.7 hours. Inference is approximately 6.3\% slower per sample (0.0101 vs. 0.0095 seconds on p3.2xlarge V100 GPUs). The increased energy consumption ($\approx$ 2.2 kWh vs. $\approx$ 2.0 kWh per full run) results in approximately 10\% higher $\text{CO}_2$ emissions ( $\approx$ 0.99 kg $\text{CO}_2$ vs. $\approx$ 0.90 kg $\text{CO}_2$), corresponding to roughly 0.09 kg additional $\text{CO}_2$ per model training. These modest computational costs are justified by the significant performance improvements as shown in Table~\ref{tab:all_results}. Detailed report is presented in Table~\ref{tab:computational_impact}.

\begin{table*}[!htbp]
	\centering
	\small
		\begin{tabular}{lcc}
			\toprule
			\textbf{Metric} & \textbf{Joint Training} & \textbf{GETR-GAT} \\
			\midrule
			\multicolumn{3}{l}{\textit{Training Configuration}} \\
			Training Instance & AWS p4de.24xlarge (8× A100 80GB) & Same \\
			Batch Size & 128 & 120 \\
			Number of Epochs & 50 & 50 \\
			\midrule
			\multicolumn{3}{l}{\textit{Training Efficiency}} \\
			Training Time per Epoch & $\approx$ 45 min & $\approx$ 50 min (+11\%) \\
			Total Training Time (50 epochs) & $\approx$ 37.5 hours & $\approx$ 41.7 hours \\
			Peak GPU Memory Usage & $\approx$ 38 GB & $\approx$ 41 GB (+8\%) \\
			\midrule
			\multicolumn{3}{l}{\textit{Inference Metrics}} \\
			Inference Instance & AWS p3.2xlarge (1× V100 16GB) & Same \\
			Inference Time per Sample & 0.0095 sec & 0.0101 sec (+6.3\%) \\
			\midrule
			\multicolumn{3}{l}{\textit{Energy \& Environmental Impact}} \\
			Energy Consumption per Run & $\approx$ 2.0 kWh &  $\approx$ 2.2 kWh (+10\%) \\
			$\text{CO}_2$ Emissions per Run & $\approx$ 0.90 kg $\text{CO}_2$ & $\approx$ 0.99 kg $\text{CO}_2$ (+10\%) \\
			\midrule
			\multicolumn{3}{l}{\textit{Performance Improvements}} \\
			Macro-F1 (Sentiment Marathi) & 0.75 & 0.87 (+12 pp) \\
			Macro-F1 (Sentiment Bangla) & 0.63 & 0.75 (+12 pp) \\
			Macro-F1 (NER Marathi) & 0.35 & 0.44 (+9 pp) \\
			Macro-F1 (NER Malayalam) & 0.28 & 0.52 (+24 pp) \\
			\bottomrule
		\end{tabular}
	\caption{Computational cost and environmental impact comparison between Joint Training and GETR-GAT. All training conducted on AWS p4de.24xlarge instances and inference on p3.2xlarge instances. Percentage increases are shown in parentheses. Despite 10--11\% overhead in training time and energy consumption, GETR-GAT achieves substantial performance improvements across all tasks.}
	\label{tab:computational_impact}
\end{table*}

\end{document}